\documentclass[11pt, a4paper, goog]{google}

\usepackage[authoryear, sort&compress, round]{natbib}
\usepackage[T1]{fontenc}
\usepackage[utf8]{inputenc}
\usepackage{titletoc}

\usepackage{multirow}
\usepackage{array}
\usepackage{tabularx}
\usepackage{pdflscape}
\usepackage{arydshln}          
\usepackage{placeins}
\usepackage{tikz}
\usetikzlibrary{arrows.meta, positioning, fit, backgrounds, calc, shapes.geometric}
\usepackage[most]{tcolorbox}   
\usepackage{soul}              
\usepackage{booktabs}
\usepackage{algorithm}
\usepackage{algpseudocode}

\usepackage{xspace}

\definecolor{datablue}{HTML}{4472C4}
\definecolor{sharedgray}{HTML}{6C757D}
\definecolor{retgreen}{HTML}{548235}
\definecolor{tvorange}{HTML}{C55A11}
\definecolor{lightbg}{HTML}{F2F2F2}
\definecolor{lightgreen}{HTML}{E2EFDA}
\definecolor{lightorange}{HTML}{FCE4CC}
\definecolor{lightblue}{HTML}{D6E4F0}
\definecolor{antgreen}{HTML}{548235}
\definecolor{conpurple}{HTML}{7030A0}
\definecolor{highcolor}{HTML}{2E75B6}
\definecolor{medcolor}{HTML}{BF8F00}
\definecolor{lowcolor}{HTML}{C00000}
\definecolor{lightpurple}{HTML}{E8D5F5}

\uselogo{} 

\newcommand{\method}{\textsc{RRSI}\xspace}

\title{\method: Regularized Recursive Self-Improvement of Agent Harnesses}

\correspondingauthor{pxia@cs.unc.edu, \{rujunh, chenyulee\}@google.com}

\author[1,2*]{Peng Xia}
\author[1]{Rujun Han}
\author[1]{Zifeng Wang}
\author[1]{Yanfei Chen}
\author[1]{Yufan Zhuang}
\author[3]{Yoonho Lee}
\author[4]{Chengsong Huang}
\author[1]{Han Yu}
\author[1]{Zhongying CuiZhu}
\author[1]{Yifei Ming}
\author[2]{Huaxiu Yao}
\author[1]{Burak Gokturk}
\author[1]{Tomas Pfister}
\author[1]{Chen-Yu Lee}

\newsavebox{\googlelogobox}
\sbox{\googlelogobox}{%
  \includegraphics[height=1.5ex]{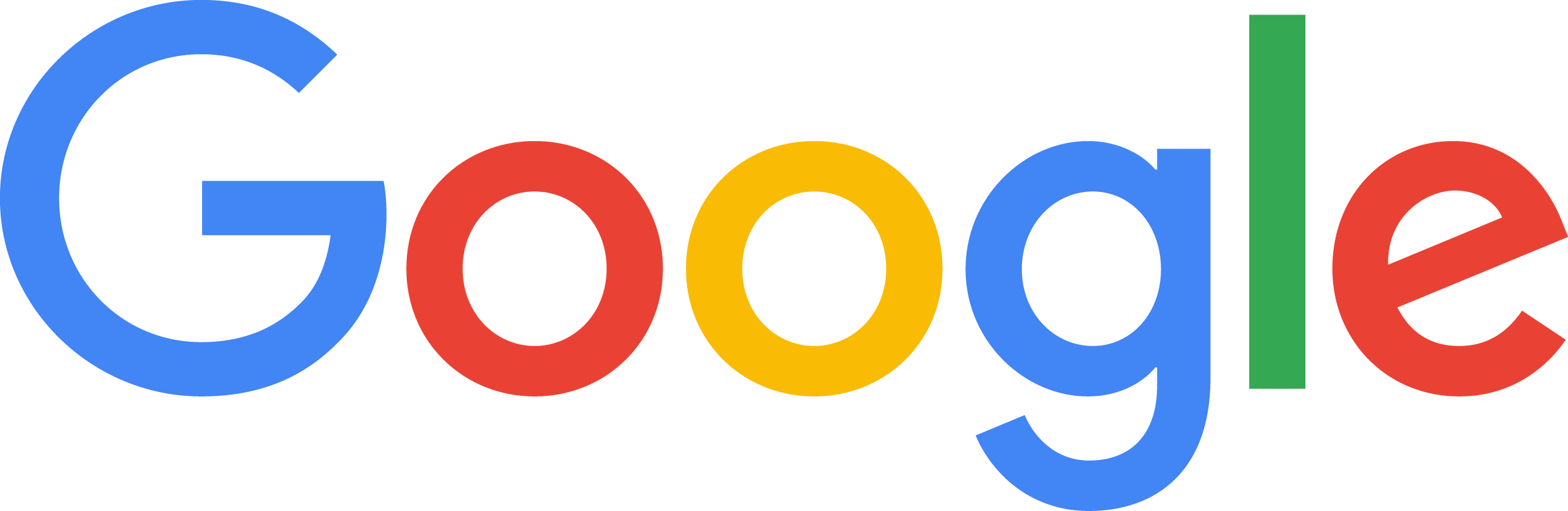}%
}

\DeclareRobustCommand{\googlelogo}{%
  \raisebox{-0.3ex}{\usebox{\googlelogobox}}%
}

\newsavebox{\unclogobox}
\sbox{\unclogobox}{%
  \includegraphics[height=1.5ex]{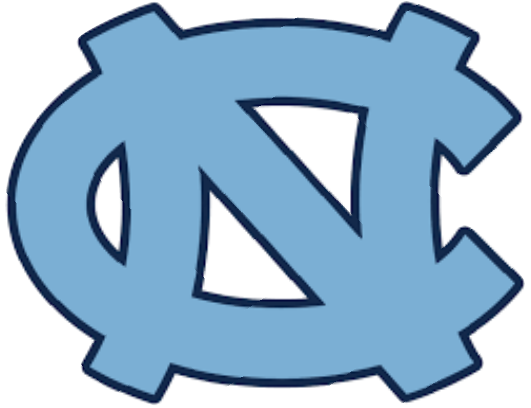}%
}

\DeclareRobustCommand{\unclogo}{%
  \raisebox{-0.3ex}{\usebox{\unclogobox}}%
}

\newsavebox{\stflogobox}
\sbox{\stflogobox}{%
  \includegraphics[height=1.5ex]{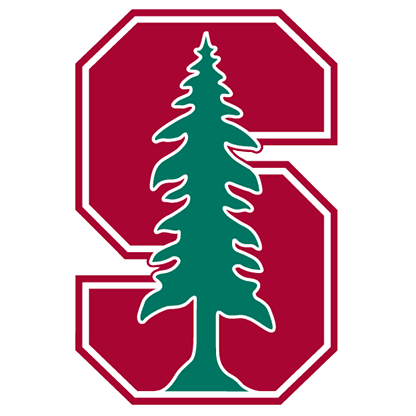}%
}

\DeclareRobustCommand{\stflogo}{%
  \raisebox{-0.3ex}{\usebox{\stflogobox}}%
}

\newsavebox{\washulogobox}
\sbox{\washulogobox}{%
  \includegraphics[height=1.5ex]{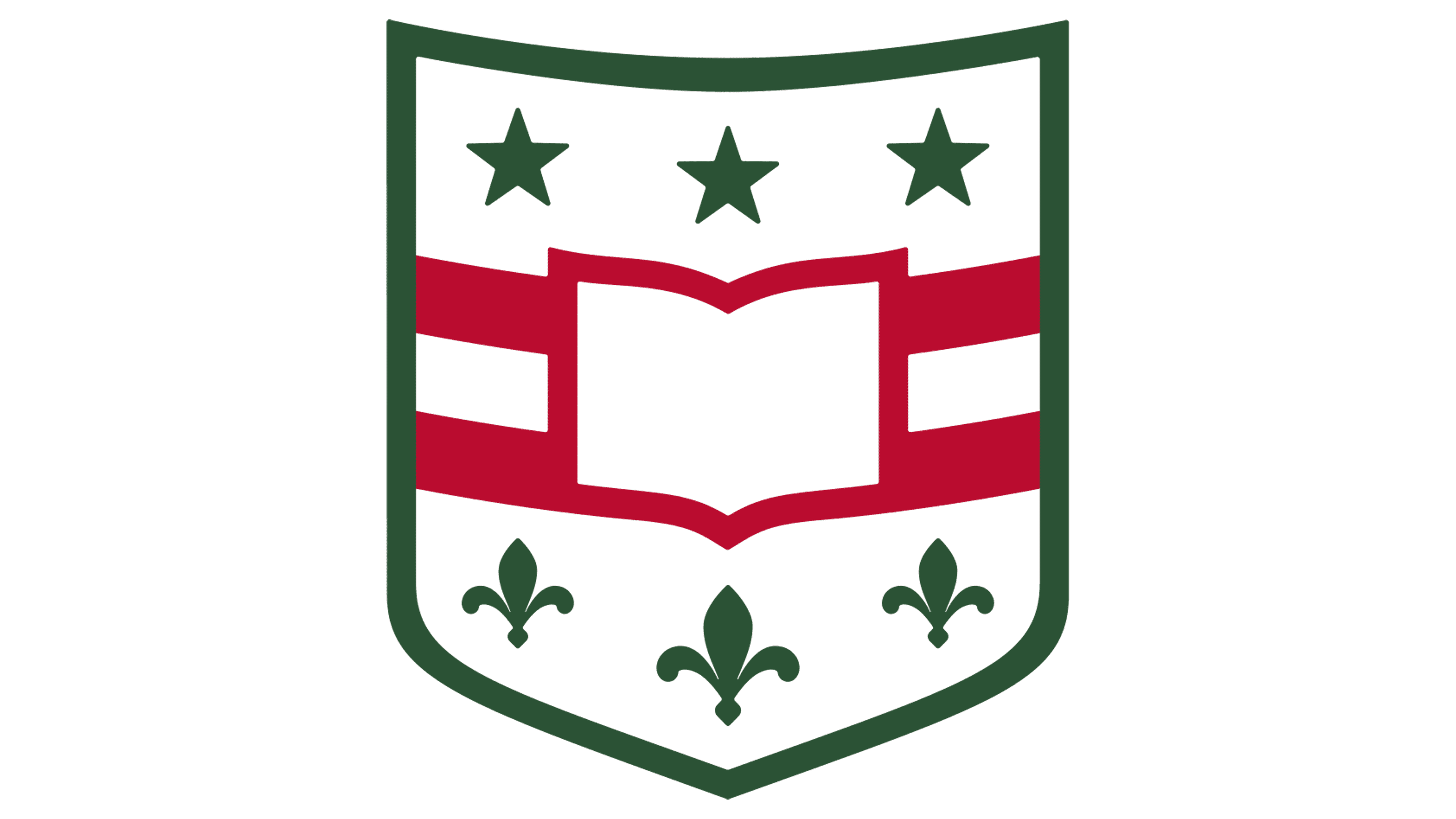}%
}

\DeclareRobustCommand{\washulogo}{%
  \raisebox{-0.3ex}{\usebox{\washulogobox}}%
}

\affil[1]{\protect\googlelogo\hspace{0.1em}  \thepa{}{} Cloud AI Research}
\affil[2]{\protect\unclogo\hspace{0.1em} UNC-Chapel Hill}
\affil[3]{\protect\stflogo\hspace{0.1em} Stanford University}
\affil[4]{\protect\washulogo Washington University in St. Louis}

\worknote{* This work was done while Peng was a Student Researcher at Google Cloud AI Research.}

\begin{abstract}
An LLM agent's capability is largely magnified by its harness, namely the prompts, control flow, tooling, memory, and context management surrounding the frozen backbone model. 
Recent methods increasingly automate this process by iteratively proposing and selecting component-wise edits of an agent harness, practically establishing a form of recursive self-improvement (RSI) at the agent-system level. However, such recursive evolution may overfit by memorizing the training tasks, showing large in-distribution gains that shrink or even vanish on out-of-distribution benchmarks. 
We introduce Regularized Recursive Self-Improvement of Agent Harnesses (\method{}), which incorporates the principles of regularizations into harness self-improvement by constraining the evolution candidate proposal and selection. 
The proposer operates with a temporally annealed budget, limiting how many edits a candidate can bundle, and it encourages unexplored trajectories based on evolution history. The selector is equipped with a critic and a pruner: the critic screens benchmark-specific proposals, while the pruner, removes changes that are too small, too expensive, or no longer useful. Together these constraints favor reusable agent mechanisms over benchmark-specific ones or even noises.
Across eight benchmarks spanning coding, agentic workspace and engineering design tasks, \method{} gains up to 6.0 points on the split it evolves against and up to 4.3 points on the five out-of-distribution benchmarks, while producing a harness that runs on 30\% fewer policy tokens than the unregularized evolution. 

\vspace{4pt}
{\centering
\raisebox{-0.16\height}{\includegraphics[height=1.05em]{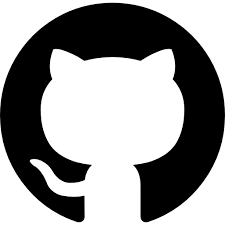}}\hspace{0.32em}\href{https://github.com/google-research/rrsi}{\texttt{github.com/google-research/rrsi}}%
\hspace{1.15em}\textcolor{gray!45}{\rule[0.02em]{0.6pt}{0.9em}}\hspace{1.15em}%
\raisebox{-0.16\height}{\includegraphics[height=1.05em]{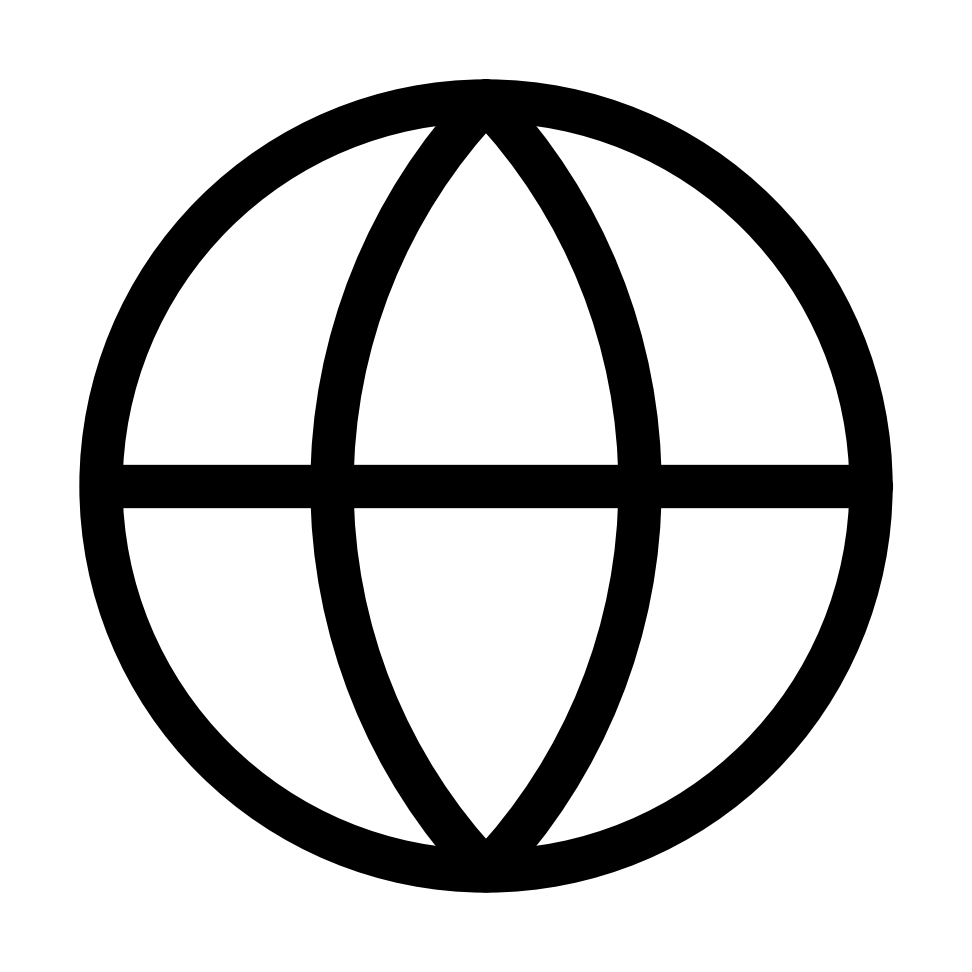}}\hspace{0.32em}\href{https://regularized-rsi.com/}{\texttt{regularized-rsi.com}}%
\par}

\end{abstract}

\begin{document}

\maketitle

\section{Introduction}
\label{sec:intro}

Modern LLM agents are systems rather than standalone models~\citep{anthropicharness,openaiharness}. A frozen backbone model is wrapped in a harness of prompts, control flow, tool interfaces, memory and context management. Agent harness decides whether the same model reads the right file before editing it, recovers from a failed command, manages efficient working context, and writes its findings into the deliverables. Much recent progress in agent products came from harness engineering rather than from new model weights~\citep{weng2026harness,zhang2026harnesses,karten2026prime}. However, this engineering relies on manual efforts, where humans inspect failed trajectories and tweak the scaffold by hand, so progress is limited by how many trajectories an engineer can read.

\begin{figure*}[t]
  \centering
  \includegraphics[width=0.95\textwidth]{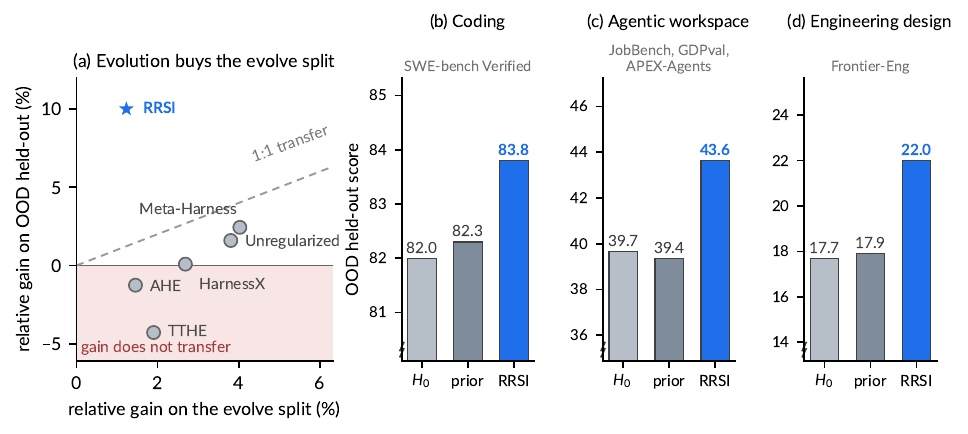}
  \vspace{-1em}
  \caption{Evolution overfits on the split it is scored on whereas \method{} generalizes the improvements. (a) Gains on the evolve split against gains out of distribution for the agentic workspace benchmark. Prior methods retain little of their evolve-set gain and several end below $H_0$, the initial harness. (b-d) Out-of-distribution held-out score for $H_0$, average of the four baseline methods and our \method{} on SWE-bench Verified, the mean of JobBench, GDPval and APEX-Agents, and Frontier-Eng. }
  \label{fig:teaser}
\end{figure*}

Recent methods automate this loop by using LLMs to optimize harness components from task feedback~\citep{lou2026autoharness,niklaus2026don,lee2026meta,lin2026agentic,zhang2026self,nie2026tthe,lee2026recursive,chen2026harnessx,karten2026continual,zhang2026darwinx}. Such iterative harness evolution provides a practical form of recursive self-improvement (RSI)~\citep{wang2025huxley,zhang2026darwin,team2026neohorse,rsi-exam-2026} at the agent-system level, where feedback from the current system is used to improve the harness that shapes its subsequent behavior. However, as illustrated in Figure~\ref{fig:teaser} (a), test-time harness evolution repeatedly proposes and selects edits using feedback from a finite evolve set, creating an adaptive overfitting risk: evolve-set performance may improve without corresponding gains on unseen tasks. Recent studies observe substantial gaps between evolution and held-out performance, and show that apparent improvements can arise from task-specific fitting or increased test-time computation rather than reusable mechanisms~\citep{wang2026rethinking,ding2026harnessdelta,lin2026harness}. Accordingly, recent works explicitly separate evolution and evaluation tasks to measure generalization~\citep{huang2026evo,ke2026evoharnessbench,zhang2026harnesscompass}. We therefore study the generalization problem in recursive self-improvement, which is defined as evolved harness transferring to unseen benchmarks with different task descriptions, tool interfaces, or verifiers.

Our study shows that overfitting can arise through several coupled behaviors~\citep{zhang2026harnesscompass,yang2026better}. The evolution search may encode benchmark-specific patterns, promote candidates favored by the evaluation noise, or accumulate complexity that improves evolve-set scores without improving the underlying agent mechanism. These \emph{benchmark-specific fitting}, \emph{noise chasing}, and \emph{complexity accumulation} all widen the evolve-to-transfer gap. Inspired by these observations, our solution regularizes how recursive harness improvements use finite and noisy feedback.

We introduce \method{}, a framework for regularizing the RSI of agent harness that keeps the harness fully editable while constraining how finite evolve-set feedback guides the search. As illustrated in Figure~2, \method{} regularizes both sides of the evolution loop: it encourages simpler and more reusable edits when proposing candidates, and applies robust selection criteria to avoid retaining improvements driven by benchmark-specific signals, evaluation noise, or unnecessary complexity. In this way, \method{} favors edits that transfer beyond evolution set without restricting which harness components may be updated.

We evaluate \method{} on eight benchmarks spanning three domains. In each domain the harness is evolved on a single suite, and is then run unchanged on held-out benchmarks. As shown in Figure~\ref{fig:teaser} (b--d), it gains up to 6.0 points on the evolving split and improves all six held-out splits, by up to 4.3 points out of distribution, on fewer policy tokens than unregularized evolution spends. More importantly, these gains generalize beyond the environment used for evolution. \method{} retains its improvements across substantially different tasks and evaluation settings, indicating that it learns broadly useful harness changes. More importantly, \method{} generalizes across held-out environments, outperforming the average prior baseline by up to 12.5\%. 

Our contributions are threefold: (1) We identify the overfitting as a key challenge in harness-based recursive self-improvement. (2) We propose \method{}, which regularizes both proposal and selection during harness evolution while keeping each harness component editable. (3) Across eight benchmarks in three domains, \method{} improves both transfer and efficiency, showing the effectiveness of our proposed approach.
\section{Preliminaries}
\label{sec:prelim}

\noindent \textbf{Agents and Harnesses.}
We consider an agent $A = (\pi, H)$ built from a backbone policy $\pi$ and a harness $H$. The harness is everything around the weights~\citep{anthropicharness,openaiharness}: the system and task prompts, the control flow that decides when the agent plans, acts, reflects or stops, the tool interfaces and their descriptions, the memory and skill files the agent may consult, and the context management that decides what the policy sees at each step. Given a task $x$ with its environment, the agent produces a trajectory $\tau \sim A(\cdot \mid x)$ and a deliverable, which a verifier scores as $r(x, \tau) \in [0, 1]$. The verifier can be a unit-test suite in coding environments or a LLM-as-a-judge program in agentic workspace environments. For a task set $\mathcal{D}$, we measure task performance and policy-token cost as
\begin{equation}
S(H; \mathcal{D}) = \mathbb{E}_{x \sim \mathcal{D}}\, \mathbb{E}_{\tau \sim A(\cdot \mid x)}[r(x,\tau)],
\qquad
C(H; \mathcal{D}) = \mathbb{E}_{x \sim \mathcal{D}}\, \mathbb{E}_{\tau \sim A(\cdot \mid x)}[c(\tau)],
\label{eq:score_cost}
\end{equation}
where $c(\tau)$ is the number of policy tokens consumed by the trajectory.

\noindent \textbf{Harness Evolution.}
Harness evolution treats $H$ as the optimization variable while keeping the backbone policy fixed~\citep{lee2026meta}. Most methods instantiate the same generic loop. At round $t$, the current harness $H_t$ is executed on an evolve set $\mathcal{D}_{\mathrm{evolve}}$ to obtain trajectories; these trajectories are summarized into feedback $\mathcal{F}_t$; a proposer LLM generates candidate harnesses; the candidates are evaluated on the same evolve set; and the best candidate is selected as the next incumbent. Abstractly,
\begin{equation}
\mathcal{H}_t = \{H_t^{(1)},\ldots,H_t^{(m_t)}\} \sim P_0(\cdot \mid H_t,\mathcal{F}_t),
\qquad
H_{t+1} = \operatorname*{arg\,max}_{H'\in \mathcal{H}_t\cup\{H_t\}} \hat S(H';\mathcal{D}_{\mathrm{evolve}}),
\label{eq:standard_evolution}
\end{equation}
where $P_0$ denotes the unconstrained proposal process and $\hat S$ is the empirical score obtained from a finite number of stochastic agent runs. With $k$ trials per task, we use
\begin{equation}
\hat{S}(H) = \frac{1}{k|\mathcal{D}_{\mathrm{evolve}}|}\sum_{x\in\mathcal{D}_{\mathrm{evolve}}}\sum_{j=1}^k r(x,\tau_x^{(j)}),
\qquad
\hat{C}(H) = \frac{1}{k|\mathcal{D}_{\mathrm{evolve}}|}\sum_{x\in\mathcal{D}_{\mathrm{evolve}}}\sum_{j=1}^k c(\tau_x^{(j)}).
\label{eq:estimate}
\end{equation}
Unlike ordinary evaluation, this reuse of $\mathcal{D}_{\mathrm{evolve}}$ is adaptive: the candidates proposed at round $t$ depend on measurements obtained from the same tasks in earlier rounds. Harness evolution can therefore be viewed as adaptive empirical optimization over an unusually expressive search space. 

\section{\method{}}
\label{sec:method}

We study RSI through iterative harness evolution, where feedback from the current agent system is repeatedly used to propose and select modifications to the harness. \method{} follows this recursive improvement process and keeps the harness edit space open, but regularizes how the evolution moves through that space. The key idea is to translate  regularization principles from machine learning into an adaptive harness search: sparse updates limit how many mechanisms can change in response to one round of feedback, evidence-aware credit assignment prevents the search from repeatedly spending its capacity on hypotheses it has already falsified, and conservative selection prevents leakage, evaluation noise, or unjustified resource growth from becoming permanent harness state.

\begin{figure}[t]
  \centering
  \includegraphics[width=0.9\textwidth]{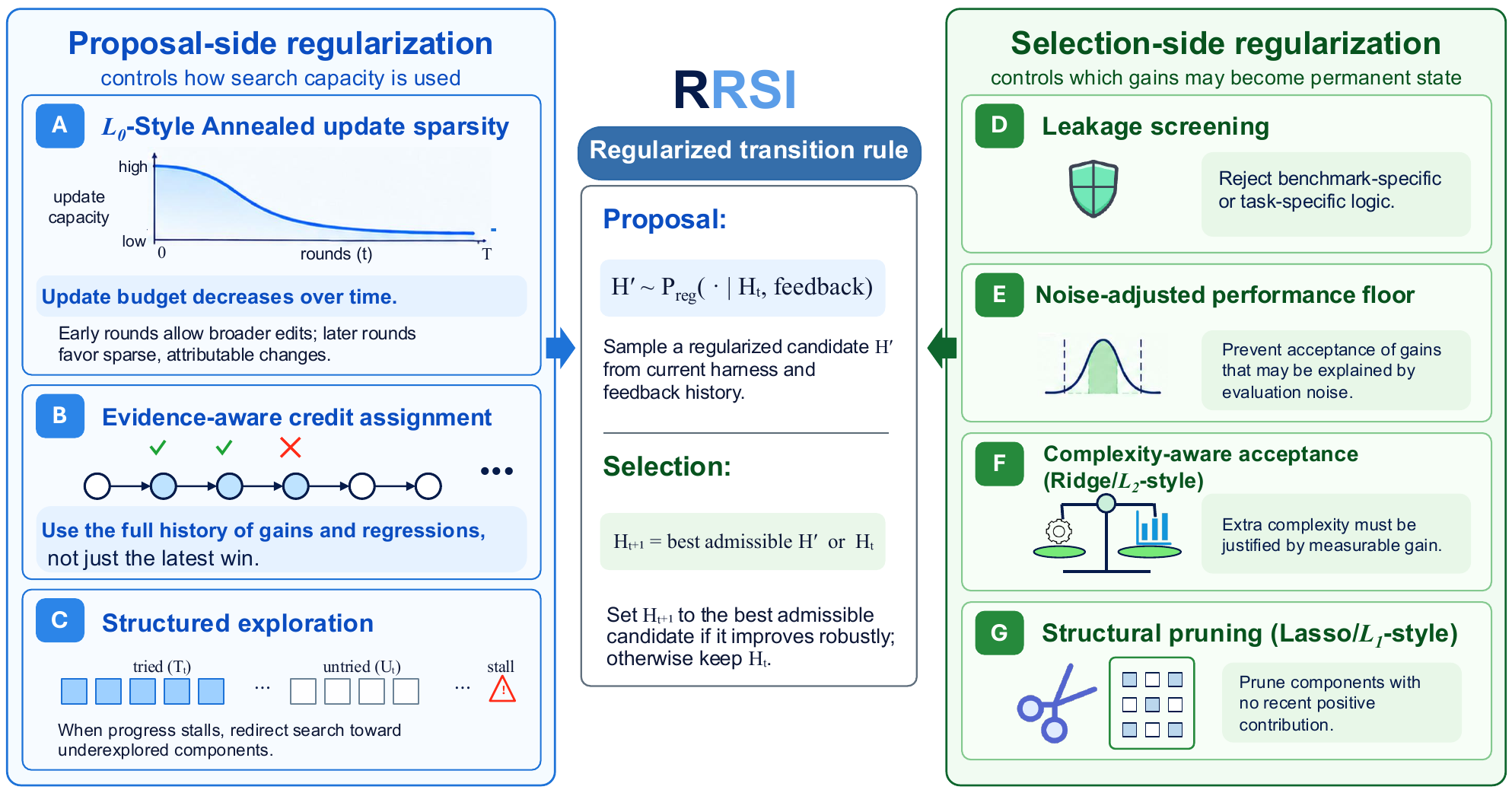}
  \caption{Overview of \method{}. \method{} regularizes the search trajectory, not restricting the potential harness edit space: proposal-side constraints control how search capacity is used, while selection-side constraints control which measured improvements are allowed to become a permanent state.}
  \label{fig:method}
\end{figure}

\subsection{A Regularization View of Harness Evolution}
\label{subsec:setup}

Let $\Omega(H)$ denote the set of harnesses reachable from $H$ by arbitrary source edits. \method{} deliberately leaves $\Omega(H)$ open: prompts, control flow, configuration, context management, tools, skills, memory, and subagents may all be modified, added, or removed. Instead of restricting this hypothesis space directly, we regularize the search trajectory through it. At each round $t$, the proposer uses feedback from the finite evolve set to generate candidate edits to the current harness $H_t$, and the selector determines which, if any, should replace the incumbent.

This view separates two complementary forms of regularization. On the proposal side, we constrain how much adaptive capacity can be exercised in a single round and where that capacity is spent. On the selection side, we constrain which empirical improvements are strong enough, efficient enough, and sufficiently free of leakage to survive. Our complexity control framework takes inspiration from three classical regularization approaches~\citep{hastie2009elements,goodfellow2016deep,louizos2018learning}, and we explain the \emph{analogy} below. The edit budget is the closest to an $L_0$-style cardinality constraint as it directly limits the number of independently active edits in an update. Structural pruning is analogous to Lasso/$L_1$-style sparsification because persistently unproductive components are removed from the retained harness, producing a sparser structure. Complexity-aware acceptance is analogous to Ridge/$L_2$-style shrinkage since it suppresses unchecked growth in the aggregate resource footprint without requiring any particular component to be eliminated. Detailed algorithm description can be found in Appendix~\ref{app:methoddetails}.

\subsection{Regularizing the Proposal Distribution}
\label{subsec:proposal}

The proposal distribution determines how aggressively the search can respond to feedback from the evolve set. \method{} regularizes it in three ways: it anneals how much update capacity a single round may exercise, it makes credit assignment evidence-aware over the whole run, and it structures where that capacity is spent.

\paragraph{$L_0$-Style Annealed Update Sparsity.}
An unconstrained proposer can bundle many unrelated modifications into one candidate. Such candidates have high effective capacity: they can fit more idiosyncrasies of the current feedback, and any measured change is difficult to attribute to a particular mechanism. We therefore cap the number of independently attributable edits that may be included in one proposal. At round $t$ of a $T$-round run, this budget is
\begin{equation}
 b_t=\Big\lceil b_{\min}+(b_{\max}-b_{\min})\cdot\tfrac{1}{2}\big(1+\cos(\pi t/T)\big)\Big\rceil.
\label{eq:anneal}
\end{equation}
The schedule decreases from $b_{\max}$ to $b_{\min}$: early rounds may combine several coordinated changes to discover new mechanisms, whereas later rounds become increasingly sparse and attributable. 
This is our most direct classical analogy: if the independently attributable edits in a candidate are represented by binary activity indicators, the budget bounds their cardinality, i.e., an $L_0$-style constraint on the update. The analogy applies to update sparsity rather than to a fixed model parameter vector; the edit pool can change across rounds, and we do not optimize an $L_0$-penalized objective.

\paragraph{Evidence-Aware Credit Assignment.}
Constraining the size of an update only helps if the search knows what earlier updates established. Every evaluation is another adaptive look at the same finite evolve set, so repeatedly testing hypotheses that earlier rounds already falsified spends search capacity without adding useful evidence~\citep{dwork2015generalization}. \method{} therefore records, for every evaluated candidate, the component it modifies, the hypothesis it tests, the source diff, the resulting score and cost changes, and whether the candidate was accepted. The proposer conditions on this history in later rounds: rejected mechanisms remain negative evidence, while successful mechanisms retain explicit credit. As later rounds allow fewer edits per candidate, it becomes easier to identify which change is responsible for an observed improvement.

\paragraph{Structured Exploration.}
The same history reveals when the proposer has collapsed onto a narrow edit family, for example repeatedly rewriting prompts while leaving agent structural mechanisms untouched. We treat the search as stalled when its progress over the previous $w$ rounds remains within the empirical noise band $\delta$. During a stall, a small portion of the proposal budget is reserved for components that have not yet been exercised in the run. This plays a role similar to diversity or entropy regularization: it redirects limited proposal capacity toward underexplored mechanisms without changing which mechanisms the harness is allowed to contain~\citep{haarnoja2018soft}.

\subsection{Regularizing Candidate Selection}
\label{subsec:acceptance}

Standard harness evolution can promote the candidate with the largest measured score even when that score reflects explicit leakage, stochastic variation, or costly growth. \method{} retains the same empirical objective but regularizes which candidates are allowed to become permanent state. A candidate must satisfy several non-compensatory criteria before its score can justify replacing the incumbent.

\paragraph{Leakage Screening.}
Before full evaluation, a critic reads each candidate diff and rejects edits that explicitly encode task names, entity names, task-specific values, answers, or other logic specific to the evolve benchmark, as well as edits that add inert machinery. The screen targets benchmark-specific content rather than particular harness components: generic prompt or tool-description improvements remain valid candidates. Screening before evaluation is important because a leaking candidate never receives the inflated evolve-set score that could make it attractive to subsequent rounds.

\paragraph{Stability-Aware Acceptance.}
Repeatedly selecting among noisy evaluations can convert stochastic winners into permanent search state. Before evolution, we repeatedly evaluate the unchanged base harness and estimate an empirical noise band $\delta$. Let $S^\star$ denote the best evolve-set score observed so far. A candidate must satisfy the noise-adjusted floor
\begin{equation}
\hat S(H')\ge S^\star-\delta.
\label{eq:floor}
\end{equation}
The floor prevents the search from walking downhill through a sequence of regressions that are individually small enough to be mistaken for noise. More broadly, it makes selection conservative to fluctuations induced by repeated stochastic evaluation on the same evolve set~\citep{dwork2015generalization}.

\paragraph{Ridge/$L_2$-Style Complexity-Aware Acceptance.}
For a candidate $H'$ relative to the current harness $H_t$, let
\begin{equation}
\Delta S=\hat S(H')-\hat S(H_t),
\qquad
\Delta C=\frac{\hat C(H')-\hat C(H_t)}{\hat C(H_t)}.
\label{eq:deltas}
\end{equation}
For a candidate whose gain exceeds the noise band, $\Delta S>\delta$, we require
\begin{equation}
\Delta C\le\beta_0+\beta_1\Delta S.
\label{eq:tokenbudget}
\end{equation}
Here, $\beta_0$ sets the cost increase tolerated for a negligible score gain, while $\beta_1$ controls how much additional cost is allowed as the measured improvement increases. We select these values on the evolve set and keep them fixed for all transfer evaluations. Thus additional inference cost must be justified by measurable performance improvement. This process is analogous to Ridge/$L_2$-style shrinkage: it discourages unconstrained growth in the overall magnitude of the solution, which is represented by the harness's aggregate resource footprint in our approach. As a shrinkage method, it does not require any particular component to be removed for sparsity. We use policy-token cost as a common measurable proxy for this footprint. This is an analogy to Ridge's non-sparsifying complexity control. The detailed rule for candidates whose measured change falls within the noise band is deferred to the Appendix~\ref{app:selection}.

\paragraph{Lasso/$L_1$-Style Structural Pruning.}
The annealed budget in Equation~\eqref{eq:anneal} sparsifies each \emph{update}; pruning sparsifies the \emph{retained harness}. \method{} tracks whether recently exercised components have produced a strictly positive measured gain over a fixed pruning window. Components that remain unproductive are reported to the proposer as deletion targets in subsequent rounds. This process imitates the Lasso/$L_1$-style sparsification: mechanisms with insufficient evidence of utility are removed entirely, so the retained harness becomes structurally sparser rather than merely cheaper in aggregate. The correspondence is again qualitative, i.e., Lasso reduces the number of parameters through $L_1$ regularization, whereas our pruning rule deletes discrete harness components based on their observed contribution. The shared intuition is selective sparsification: a mechanism must continue to earn its place rather than persist simply because score-only evolution has no incentive to remove it~\citep{hastie2009elements}.

\section{Experiments}

We evaluate \method{} on eight benchmarks spanning three domains: Terminal-Bench 2.1 and SWE-bench Verified for coding, Harvey LAB, JobBench, GDPval and APEX-Agents for agentic workspace tasks, and EngDesign and Frontier-Eng for engineering design. Our experiments address the following questions: 1) How does \method{} compare to state-of-the-art harness evolution methods? 2) Do the gains transfer to in-distribution held-out tasks and to out-of-distribution benchmarks that the search never saw? 3) What is the contribution of different components? 4) How does the harness evolve over a run, and what does it cost in tokens?

\subsection{Experimental Setup}

\noindent \textbf{Environments.}
We evolve harnesses in three types of tasks, i.e., coding tasks, agentic workspace tasks and engineering design tasks. For coding, Terminal-Bench 2.1~\citep{merrill2026terminal} is a suite of 89 containerized terminal tasks in which the agent drives a real shell and is verified by the task's own unit tests. For agentic workspace tasks, Harvey LAB~\citep{harveylab2026} is a legal-work benchmark spanning 25 practice areas. It is split into a fixed evolve set of 120 tasks and a pristine in-distribution held-out set of 40 tasks. For engineering design, EngDesign~\citep{guo2025toward} contributes 61 design tasks, each graded by its own frozen simulator rather than by a judge model. To test its generalization capability, we additionally evaluate on out-of-distribution (OOD) held-out benchmarks: SWE-bench Verified~\citep{jimenez2024swe} for repository-level bug fixing on the coding task, JobBench~\citep{li2026jobbench}, GDPval~\citep{patwardhan2026gdpval} and APEX-Agents~\citep{vidgen2026apex} on the agentic workspace task, and Frontier-Eng~\citep{chi2026frontier} on the engineering design task.

\noindent \textbf{Baselines.}
We compare against the unevolved base harness $H_0$ that every run starts from, and against four recent harness evolution methods, Meta-Harness~\citep{lee2026meta}, AHE~\citep{lin2026agentic}, TTHE~\citep{nie2026tthe} and HarnessX~\citep{chen2026harnessx}. All the baselines start from the same $H_0$ and share the frozen policy, the evolve set and the candidate budget. The detailed descriptions of baselines are given in Appendix~\ref{app:baselines}.

\noindent \textbf{Implementation Details.}
The policy is frozen throughout Claude Opus 4.8~\citep{claudeopus48} across all three domains. The proposer, the analyst that writes the cross-round failure feedback and the leakage critic are all Claude Opus 4.8. The base harness we used are Terminus-2 (for coding)~\citep{merrill2026terminal}, a ReAct loop~\citep{yao2022react} over an MCP tool gateway, a dynamic toolbelt~\citep{vidgen2026apex}, and ReSum-style context management (for Harvey LAB and EngDesign). Further hyperparameters are given in Appendix~\ref{app:hypselect}.

\subsection{Main Results}
\label{subsec:main}
\begin{figure}[t]
  \centering
  \includegraphics[width=0.92\textwidth]{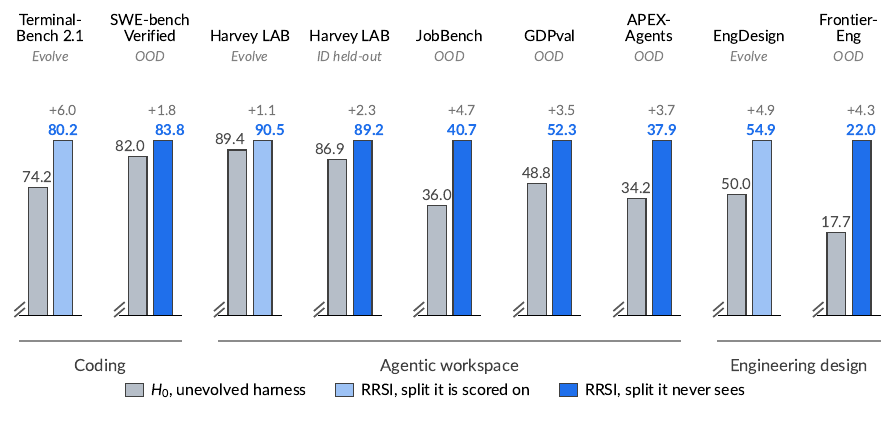}
  \caption{Main results in all three domains. }
  \label{fig:main}
\end{figure}

\noindent \textbf{\method{} improves every split outside the evolve set, in all three domains.}
Figure~\ref{fig:main} reports every number against the unevolved base harness $H_0$ measured in the same window, so no gain can be attributed to drift in the evaluation infrastructure. The evolve-set gains are 6.0 points on Terminal-Bench 2.1, 4.9 on EngDesign and 1.1 on Harvey LAB. What matters is what remains once the harness leaves those splits. SWE-bench Verified gains 1.8 points although repository-level bug fixing was never scored. The in-distribution held-out split of Harvey LAB gains 2.3, and the three out-of-distribution agentic benchmarks gain between 3.5 and 4.7 points, 7.2\% to 13.1\%. Frontier-Eng gains 4.3 Medal points, a 24.3\% relative improvement. No held-out split regresses anywhere, which is the failure a memorizing harness produces.

\begin{table}[t]
\centering
\small
\resizebox{\textwidth}{!}{
\begin{tabular}{lccccc}
\toprule
& \multicolumn{2}{c}{In-Distribution} & \multicolumn{3}{c}{Out-of-Distribution} \\
\cmidrule(lr){2-3} \cmidrule(lr){4-6}
Method & Harvey LAB (Evolve) & Harvey LAB (ID Held-out) & JobBench & GDPval & APEX-Agents \\
\midrule
$H_0$ (no evolution)            & 89.4 & 86.9 & 36.0 & 48.8 & 34.2 \\
\midrule
Meta-Harness~\citep{lee2026meta}   & \textbf{93.0} & 89.2 & 37.1 & 49.1 & 35.7 \\
AHE~\citep{lin2026agentic}         & 90.7 & 88.7 & 37.2 & 47.2 & 33.1 \\
TTHE~\citep{nie2026tthe}           & 91.1 & 88.5 & 35.2 & 47.0 & 31.7 \\
HarnessX~\citep{chen2026harnessx}   & 91.8 & 89.1 & 36.3 & 48.5 & 34.3 \\
\midrule
\method{} (ours)                & 90.5 & \textbf{89.2} & \textbf{40.7} & \textbf{52.3} & \textbf{37.9} \\
\bottomrule
\end{tabular}
}
\caption{Comparison with prior harness evolution methods on agentic workspace tasks.}
\label{tab:baselines}
\end{table}

\noindent \textbf{\method{} consistently outperforms baselines on all held-out datasets.}
Table~\ref{tab:baselines} runs the four prior methods from the same $H_0$ on the same evolve split under the same candidate budget. Every one of them works well on evolve set. The performance on in-distribution held-out split is quite similar. The separation appears out of distribution, and there the ranking inverts. Meta-Harness, the strongest baseline on the evolve split, adds 0.9 points to the out-of-distribution average; HarnessX lands on the base one; AHE and TTHE finish below the harness they started from, TTHE by 1.7 points. \method{} posts the smallest evolve-set gain of any evolved harness and the only out-of-distribution average that clears $H_0$ by more than a point, 43.6 against 39.7, which is the trade the regularizers are designed to make.

\noindent \textbf{The transfer is not an artifact of judge-mediated grading or of a shared task format.}
Harvey LAB, JobBench and GDPval are all scored by a judging model, so a harness could in principle raise its score by writing the way a judge rewards rather than by producing better work. The engineering design instance closes that route: each EngDesign and Frontier-Eng task is graded by its own simulation or testbench, the grading is deterministic, and a design either meets the stated constraints or does not. The gains survive there unchanged, and deterministic grading also removes judge variance from the measurement.

\subsection{Analysis}
\label{subsec:analysis}

The main results establish that the evolved harnesses transfer; this section asks what produced that property, whether it depends on the backbone the search was run with, and what it costs. Unless stated otherwise, every run below uses the agentic workspace instance and shares the base harness, policy, evolve split, round count and candidate budget of the main experiment, so that arms differ only in the factor under study.

\begin{table}[t]
\centering
\small
\resizebox{\textwidth}{!}{
\begin{tabular}{lcccc}
\toprule
Variant & Harvey LAB (Evolve) & Harvey LAB (ID Held-out) & OOD Avg.\ & Tokens/trial (m) $\downarrow$ \\
\midrule
$H_0$ (no evolution)                       & 89.4 & 86.9 & 39.7 & 1.56 \\
Unregularized evolution                    & 92.8   & 88.9   & 40.3   & 3.80 \\
\midrule
w/o proposal regularizers & 90.7 & 88.8 & 41.9 & 2.69 \\
w/o acceptance regularizers & 91.5 & 88.7 & 41.0 & 3.59 \\
\method{}                           & 90.5 & 89.2 & 43.6 & 2.42 \\
\bottomrule
\end{tabular}
}
\caption{Ablation study of the regularizers on agentic workspace tasks. OOD Avg.\ is the mean over JobBench, GDPval and APEX-Agents.}
\label{tab:ablation}
\end{table}

\noindent \textbf{Ablation Analysis.}
We ablate the two groups of regularizers, (i) the proposal-side constraints and (ii) the acceptance-side constraints. As shown in Table~\ref{tab:ablation}, removing either group raises the evolve-set score and lowers transfer. Without the acceptance constraints the evolve-set score rises from 90.5 to 91.5 while the out-of-distribution average falls from 43.6 to 41.0 and token cost rises by half, showing that an unconstrained selection rule spends most of its accepted edits on noise and on context rather than on mechanism. Removing the proposal constraints costs only 0.2 points on the evolve split but 1.7 out of distribution, suggesting that steering where the search looks matters even when nothing is rejected. The most significant degradation comes from removing both, which lifts the evolve-set score to 92.8, the highest of any arm, and leaves the out-of-distribution average at 40.3, within a point of the unevolved harness, at 3.80 million tokens per trial against our 2.42.

\begin{table}[t]
\centering
\small
\begin{tabular}{llccc}
\toprule
Policy & Benchmark & $H_0$ & \method{} & $\Delta$ \\
\midrule
\multirow{2}{*}{Claude Opus 4.8}
    & Terminal-Bench 2.1 (Evolve) & 74.2 & \textbf{80.2} & +6.0 \\
    & SWE-bench Verified (OOD)    & 82.0 & \textbf{83.8} & +1.8 \\
\midrule
\multirow{2}{*}{Gemini 3.5 Flash}
    & Terminal-Bench 2.1 (Evolve) & 64.6 & \textbf{78.7} & +14.1 \\
    & SWE-bench Verified (OOD)    & 76.8 & \textbf{79.0} & +2.2 \\
\bottomrule
\end{tabular}
\caption{Policy robustness in the coding domain. Harness evolution is run independently with each frozen policy on Terminal-Bench, and the harness is evaluated unchanged on SWE-bench Verified.}
\label{tab:crosspolicy}
\end{table}

\noindent \textbf{\method{} is not tied to one policy family.}
To test whether the gains from regularized harness evolution depend on the policy used during search, we independently run the coding evolution with two policy models from different families: Claude Opus 4.8 and Gemini 3.5 Flash~\citep{gemini35flash}. For each policy, we start from the same coding harness, evolve only on Terminal-Bench 2.1, and evaluate the resulting harness on both the evolve benchmark and SWE-bench Verified. As shown in Table~\ref{tab:crosspolicy}, under Gemini 3.5 Flash, \method{} improves Terminal-Bench 2.1 from 64.6 to 78.7 and transfers a 2.2-point gain to SWE-bench Verified. Under Claude Opus 4.8 the pattern is the same: Terminal-Bench 2.1 rises from 74.2 to 80.2 and SWE-bench Verified from 82.0 to 83.8, although the stronger policy starts closer to the ceiling of both suites and leaves less room to gain. In both cases the harness improves the unseen benchmark without ever being scored on it, which suggests that the benefits of \method{} are not specific to a particular backbone. 

\noindent \textbf{The evolved harness still helps under a backbone the search never used.}
A harness is a program, not a set of weights, so a mechanism that helps only the policy it was searched against is an artifact of that policy rather than a reusable one. We take the final harness of the coding run, evolved with Gemini 3.5 Flash, and evaluate it unchanged with Gemini 3.1 Flash Lite, a smaller model that never took part in the search. As shown in Table~\ref{tab:crossmodel}, Terminal-Bench 2.1 accuracy rises from 11.2 to 14.6, a 30.4\% relative gain against a base score less than a fifth of the search policy's. The mechanisms therefore do not depend on the capability level they were searched at, although the absolute gain is smaller because a weaker backbone leaves fewer tasks within reach of any harness.

\begin{table}[t]
\centering
\small
\begin{tabular}{lccc}
\toprule
Evaluation policy & $H_0$ & \method{} & $\Delta$ \\
\midrule
Gemini 3.5 Flash (search policy)    & 64.6 & \textbf{78.7} & +14.1 \\
Gemini 3.1 Flash Lite (unseen)  & 11.2 & \textbf{14.6} & +3.4 \\
\bottomrule
\end{tabular}
\caption{Cross-model transfer on Terminal-Bench 2.1. The harness evolved with Gemini 3.5 Flash as the frozen policy is run unchanged with a weaker backbone that never took part in the search.}
\label{tab:crossmodel}
\end{table}

\begin{figure}[t]
  \centering
  \includegraphics[width=0.9\textwidth]{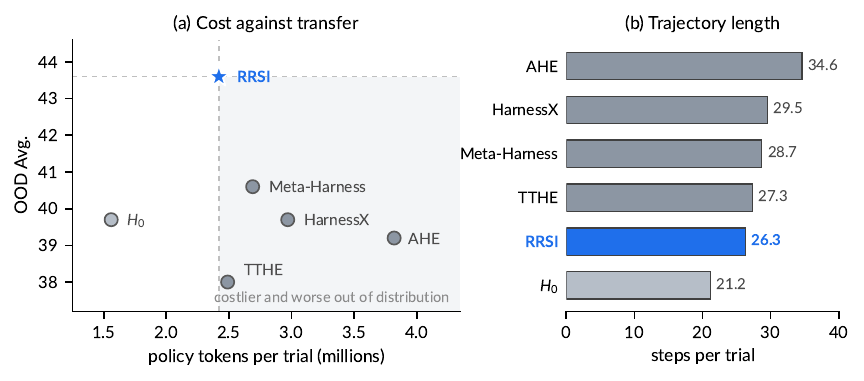}
  \caption{Cost of the final harness of each arm, measured on the evolve split of the agentic workspace instance. OOD Avg.\ is the mean over JobBench, GDPval and APEX-Agents. The shaded region in (a) is everything \method{} dominates: more policy tokens per trial for a lower out-of-distribution average.}
  \label{fig:efficiency}
\end{figure}

\noindent \textbf{\method{} produces the lightest harness of any evolved harness.}
Two regularizers act directly on cost: the $L_1$-style budget refuses growth that is not paid for when it is proposed, and the pruning rule removes growth that has stopped being paid for since. No prior method carries either constraint, and Figure~\ref{fig:efficiency} (a) shows the consequence: all four sit in the region \method{} dominates, spending more policy tokens per trial for a lower out-of-distribution average. AHE is the extreme case, at 3.82 million tokens per trial, 58\% more than ours, for 4.4 points less out of distribution. The ordering carries over to trajectory length in Figure~\ref{fig:efficiency} (b), where \method{} runs 26.3 steps per trial against 27.3 to 34.6 for the prior methods. No evolved harness is as cheap as $H_0$, at 1.56 million tokens and 21.2 steps, so evolution does buy part of its gain with test-time compute; the budget decides how much.

\section{Related Work}
\label{sec:related}

\noindent \textbf{Agent harnesses.}
The harness, and not only the backbone model, determines what an agent can accomplish: engineering reports from frontier labs describe how prompt structure, tool interfaces, context compaction and recovery logic decide whether a long-running agent finishes a task at all~\citep{anthropicharness,openaiharness}, and recent analyses argue that harnesses compose and generalize in their own right~\citep{zhang2026harnesses,weng2026harness,wang2026harness}. A well-designed harness can even substitute for scale, recovering much of a larger backbone's capability at a fraction of the cost~\citep{yang2026better}. This engineering is overwhelmingly manual, and because the best harness is tied to a specific backbone, its cost is paid again with every model release~\citep{huang2026envharness}.

\noindent \textbf{Harness evolution.}
The closest line of work automates that loop: an LLM proposer rewrites the harness and edits are kept if they raise a benchmark score~\citep{lee2026meta,lin2026agentic,nie2026tthe,chen2026harnessx,zhang2026self,lee2026recursive,karten2026continual,liu2026adaptive}, or a single component is evolved, such as skills~\citep{yang2026skillopt,xia2026skillrl}, memory~\citep{tang2025agent,ouyang2026reasoningbank,liu2026evolvemem,wu2026automem} or a preference signal over rollouts~\citep{pan2026retrospective}. This inherits both the mechanisms and the risks of self-improving agents that search over their own code under an empirical fitness signal~\citep{zhang2026darwin,wang2025huxley,zhang2026hyperagents,xia2025agent0,xia2026metaclaw,huang2026r,huang2026g}. Throughout, the search is driven by the score on the suite it optimizes against, with no term for generalization, and the cost is not hypothetical: reported gains often do not survive a change of suite~\citep{wang2026rethinking,huang2026evo}, and delta attribution separates edits that install a reusable mechanism from those that merely fit the evolution tasks~\citep{ding2026harnessdelta}. Concurrent work also targets generalization directly, either as an explicit objective of the search~\citep{zhang2026harnesscompass} or by replacing greedy selection with a diversity-preserving archive over candidate harnesses~\citep{luo2026self}. Our contribution is orthogonal to what these methods edit. We keep the same open edit space and instead regularize the search dynamics: credit assigned over the full evolution history, task-specific logic filtered before scoring, and acceptance against a noise-adjusted baseline, so what survives is a mechanism rather than a fit to the evolution suite.
\section{Conclusion}
\label{sec:conclusion}

We study iterative harness evolution as a practical form of recursive self-improvement at the agent-system level, and show that this recursive process itself requires regularization. Because a finite evolve set is reused adaptively across rounds, apparent self-improvement can reflect benchmark-specific fitting, evaluation noise, or unnecessary complexity rather than transferable progress. \method{} addresses this problem by regularizing both proposal and selection while leaving the harness edit space open. Across coding, agentic workspace, and engineering design tasks, the resulting harnesses improve held-out and cross-benchmark performance while using less inference cost than unregularized evolution. These results suggest that making agent systems increasingly capable through recursive self-improvement requires controlling not only what can change, but also how repeated feedback is converted into persistent changes.

\section*{Limitations}
Our study focuses on harness-level recursive self-improvement with frozen backbone models, and therefore does not address settings where model weights are updated during evolution. In addition, \method{} still relies on a finite evolve set and several regularization hyperparameters, so its effectiveness may depend on the quality of the feedback signal and the chosen search budget. Finally, although we evaluate transfer across multiple domains, benchmarks, and policy models, broader validation is needed to determine how well the method generalizes to substantially different agent architectures, tool ecosystems, and longer-running self-improvement processes.

\bibliography{main}

\clearpage
\appendix


\vspace*{1em} 
\noindent{\Large \textbf{Contents of Appendix}} 
\vspace{0.5em}
\hrule height 0.8pt 
\vspace{1em} 

\startcontents[appendix]
\printcontents[appendix]{l}{1}{\setcounter{tocdepth}{2}}

\clearpage

\section{Limitations}
Our study focuses on harness-level recursive self-improvement with frozen backbone models, and therefore does not address settings where model weights are updated during evolution. In addition, \method{} still relies on a finite evolve set and several regularization hyperparameters, so its effectiveness may depend on the quality of the feedback signal and the chosen search budget. Finally, although we evaluate transfer across multiple domains, benchmarks, and policy models, broader validation is needed to determine how well the method generalizes to substantially different agent architectures, tool ecosystems, and longer-running self-improvement processes.

\section{Evaluation}
\label{app:eval}

This shows how each environment is run and scored. A harness and its baseline are always evaluated in the same window, with the same tool environment, the same judge and the same number of trials.

\subsection{Terminal-Bench 2.1}
Each task is a container image with a task description, a working directory and a set of unit tests that are hidden from the agent~\citep{merrill2026terminal}. The agent drives a real shell through the harness, and a task counts as solved only if the task's own test suite passes after the agent stops, so the reward is exact and cannot be produced by a plausible-looking answer. The reported accuracy is the fraction of the 89 tasks solved in this way. Containers are torn down and rebuilt between arms so that no state carries from one evaluation to the next.

\subsection{SWE-bench Verified}
Each instance is a real GitHub issue paired with the repository snapshot at the time of the report~\citep{jimenez2024swe}. The agent must produce a patch, which is then applied to the snapshot and checked against the instance's fail-to-pass tests, which must go from failing to passing, and its pass-to-pass tests, which must remain passing. The reported resolve rate is the fraction of instances that satisfy both conditions.

\subsection{Harvey LAB}
Each task provides a folder of source documents in Word, Excel and PDF form and requires the agent to produce deliverable files under exact requested filenames~\citep{harveylab2026}, which are graded by a strict per-criterion rubric of 20 to 100 independently judged criteria per task, roughly 14{,}000 criterion verdicts per full evaluation. A criterion is judged in isolation by an LLM judge (Gemini-3.5-Flash~\citep{gemini35flash}) that reads the produced deliverable together with that single criterion, and the score of a run is the fraction of criteria passed over all tasks, so a task with a long rubric contributes proportionally more evidence than a short one and a missing deliverable fails every criterion it was supposed to satisfy rather than being dropped. The 160 tasks are partitioned once into a 120-task evolve set and a 40-task held-out set, and the partition is fixed for the experiment. 

\subsection{JobBench}
Tasks are drawn from real professional workflows~\citep{li2026jobbench}, each shipping a task folder of input files and a wrapper prompt, with the reference material the agent would need to look up deliberately withheld so that part of the work is genuine retrieval. The harness exposes a filesystem, a code execution tool for producing office and PDF deliverables, and a grounded web search tool. Deliverables are graded by the benchmark's own weighted rubric, and the reported number is the weighted rubric score over the evaluated split. We use an LLM judge (average score of Gemini-3.5-Flash and Claude Opus 4.8).

\subsection{GDPval}
For each task the deliverable produced by the harness is placed side by side with the human expert deliverable shipped with the benchmark~\citep{patwardhan2026gdpval}, a panel of three judges of different provenance picks the better of the two, and the reported number is the win rate against the expert over 185 tasks. The panel combines an open-weight model served locally (Qwen3.6-35B-A3B~\citep{qwen36plus}) with two proprietary models from different vendors (Claude Sonnet 4.6~\citep{claudesonnet46} and Gemini-3.1 Pro~\citep{gemini31pro}), each pair is judged in both presentation orders to remove position bias, and the verdict for a task is the majority vote of the three. Each judge therefore issues 204 comparisons per harness, and a win rate above 50\% means the harness produces the preferred deliverable more often than the human expert it is compared against.

\subsection{APEX-Agents}
Each task places the agent in a sandboxed world with its own MCP tool surface, covering a filesystem, PDF reading, spreadsheets, mail, chat, calendar, documents and code execution, and spanning three professional domains~\citep{vidgen2026apex}. A task is graded by a per-task rubric judged by an LLM judge (Gemini-3.5-Flash), and a task counts as a success under pass@1 only when its rubric is satisfied on the single sampled rollout. We evaluate the full set of 480 tasks and always report over that full denominator, so a task whose rollout is missing because of an infrastructure failure counts as a failure rather than being excluded, which prevents a harness that crashes on hard worlds from looking better than one that attempts them.

\subsection{EngDesign}
We used the license-free subset of EngDesign~\citep{guo2025toward}, of which we take the 61 tasks that run without proprietary simulators. Each task states a design goal together with the physical constraints the design must satisfy, and each is graded by its own frozen simulation or testbench rather than by a judge model, so grading is deterministic and every point of variance we measure comes from the policy. Evolution runs on all 61 tasks with no in-distribution held-out split, since the suite is too small to spend tasks on one.

\subsection{Frontier-Eng}
Frontier-Eng~\citep{chi2026frontier} collects real-world engineering optimization problems from 26 domains. Each task asks the agent to produce a design or a program that is scored by a frozen task-specific simulator or evaluator on a continuous objective, so as with EngDesign no judge model is involved and grading is deterministic. Because the objectives are not commensurable across tasks, the benchmark reports a Medal Score: for each task the three best feasible results of the frozen v1 snapshot are the gold, silver and bronze thresholds, a submission earns 1, 0.67 or 0.33 for reaching each, and the score is the mean credit over the 47 tasks of the v1 set, which we report as a percentage. We use Frontier-Eng only as an out-of-distribution test surface. Its EngDesign domain reuses tasks from our evolve set and is excluded, and tasks whose evaluation environment could not be built in our sandbox receive no credit in either arm, so 38 of the 47 tasks contribute credit and both arms are scored on exactly the same tasks.

\section{Baseline Methods}
\label{app:baselines}

We briefly summarize the four harness-evolution baselines used in our experiments.

\noindent \textbf{Meta-Harness}~\citep{lee2026meta}.
Meta-Harness formulates harness engineering as an outer-loop optimization problem over executable harness code. Its agentic proposer has access to the source code, evaluation scores, and execution traces of previous candidates, and uses this accumulated experience to propose improved harnesses.

\noindent \textbf{Agentic Harness Engineering (AHE)}~\citep{lin2026agentic}.
AHE uses an observability-driven evolution loop for coding-agent harnesses. It organizes harness components, execution experience, and edit outcomes into explicit representations so that an evolving agent can diagnose failures, propose changes, and evaluate the effects of previous edits.

\noindent \textbf{Test-Time Harness Evolution (TTHE)}~\citep{nie2026tthe}.
TTHE evolves executable harnesses during test-time adaptation while keeping the underlying model weights fixed. It maintains multiple candidate harnesses, proposes modifications from execution traces, and uses an agentic judge to select a harness that persists to subsequent inputs.

\noindent \textbf{HarnessX}~\citep{chen2026harnessx}.
HarnessX represents an agent harness as a composition of modular, typed primitives spanning components such as prompts, tools, memory, and control flow. Its trace-driven adaptation mechanism uses execution feedback to modify and select harness configurations, enabling the runtime scaffold to evolve over time.

\section{Method Details}
\label{app:methoddetails}

This section gives the round-level formulation and implementation details omitted from Section~\ref{sec:method}. It specifies the same proposal- and selection-side regularizers used in the experiments. As in the main text, the $L_0$, Lasso/$L_1$, and Ridge/$L_2$ terminology is used only to indicate \emph{analogous roles} in complexity control. The procedure does not optimize the corresponding norm-penalized objectives, and heterogeneous harness components are not treated as coordinates of a shared continuous parameter vector.

\subsection{Round-Level Formulation}

Let $\Omega(H)$ denote the set of harnesses reachable from $H$ by arbitrary source edits. \method{} leaves $\Omega(H)$ open and instead regularizes the transition through this space. A round takes the form
\begin{equation}
\mathcal{H}_t \sim P_{\mathrm{reg}}\big(\cdot \mid H_t,\mathcal{F}_t,\mathcal{L}_t,b_t,\mathcal{E}_t,\mathcal{B}_t\big) \subseteq \Omega(H_t),
\qquad
H_{t+1}=\operatorname*{arg\,max}_{H'\in\mathcal{H}_t\cap\mathcal{A}_t}\hat S(H'),
\label{eq:transition}
\end{equation}
with $H_{t+1}=H_t$ if no candidate is admissible. Here $\mathcal{F}_t$ is feedback from the current round, $\mathcal{L}_t$ is the edit history, $b_t$ is the annealed edit budget from Equation~\eqref{eq:anneal}, $\mathcal{E}_t$ contains exploration directives, $\mathcal{B}_t$ contains structural pruning targets inferred from recent history, and $\mathcal{A}_t$ is the set of candidates allowed to replace the incumbent.

A run applies Algorithms~\ref{alg:propose} and~\ref{alg:accept} for $t=0,\ldots,T-1$, starting from $H_0$ with $S^\star=\hat S(H_0)$. Before evolution, the unchanged base harness is evaluated repeatedly to estimate the empirical noise tolerance $\delta$.

\begin{algorithm}[t]
\footnotesize
\caption{\method{}, proposal side: one round of regularized candidate generation.}
\label{alg:propose}
\begin{algorithmic}[1]
\Require incumbent $H_t$, edit history $\mathcal{L}_t$, round $t$ of horizon $T$
\Require budgets $b_{\min},b_{\max}$, stall window $w$, noise band $\delta$, exploration slots $m_{\mathrm{draft}}$
\State $\mathcal{F}_t\gets\textsc{Analyze}(H_t,\mathcal{D}_{\mathrm{evolve}})$ \Comment{failure feedback of the incumbent}
\State $b_t\gets\big\lceil b_{\min}+(b_{\max}-b_{\min})\tfrac{1}{2}(1+\cos\tfrac{\pi t}{T})\big\rceil$ \Comment{Eq.~\eqref{eq:anneal}: annealed budget}
\State $\sigma_t\gets\mathbb{1}[\hat S_t-\hat S_{t-w}\le\delta]$ \Comment{stalled if $w$ rounds show no improvements}
\State $\mathcal{T}_t\gets\{\ell_i:(t_i,\ell_i,\ldots)\in\mathcal{L}_t\}$ \Comment{log components already explored}
\State $\mathcal{U}_t\gets\mathcal{K}\setminus\mathcal{T}_t$ \Comment{log components never explored}
\State $\mathcal{E}_t\gets(\sigma_t,\mathcal{U}_t,m_{\mathrm{draft}})$ \Comment{Eq.~\eqref{eq:explore}: reserve slots when stalled}
\State $\mathcal{B}_t\gets\{\ell\in\mathcal{T}_t:g_t(\ell)\le0\}$ \Comment{Eq.~\eqref{eq:prune}: delete the unproductive}
\State $\mathcal{H}_t\sim P_{\mathrm{reg}}(\cdot\mid H_t,\mathcal{F}_t,\mathcal{L}_t,b_t,\mathcal{E}_t,\mathcal{B}_t)$ \Comment{perform regularized edits}
\State tag each atomic edit with its component and hypothesis \Comment{for attributable later edits}
\State \Return candidates that pass the pre-evaluation screen \Comment{the critic reads the diff, not the proposal}
\end{algorithmic}
\end{algorithm}

\begin{algorithm}[t]
\footnotesize
\caption{\method{}, selection side: which of the round's candidates may replace the incumbent.}
\label{alg:accept}
\begin{algorithmic}[1]
\Require screened candidates $\mathcal{H}_t$, incumbent $(H_t,\hat S_t,\hat C_t)$, best-ever $S^\star$, band $\delta$, trials $k$
\Require cost allowances $\beta_0,\beta_1$, within-band weights $w_s,w_c,w_n$
\State $\mathcal{A}_t\gets\emptyset$ \Comment{candidates admissible this round}
\For{$H'\in\mathcal{H}_t$ \textbf{in parallel}}
  \State $\hat S',\hat C'\gets\textsc{Evaluate}(H',\mathcal{D}_{\mathrm{evolve}},k)$ \Comment{eval on the full evolve set}
  \State $\Delta S\gets\hat S'-\hat S_t$; \ $\Delta C\gets(\hat C'-\hat C_t)/\hat C_t$ \Comment{check score gain and token growth}
  \State $\nu\gets\nu_t(H')$ \Comment{current new component edit}
  \If{$\Delta S>\delta$} \Comment{the gain clears the noise}
    \State $c\gets[\Delta C\le\beta_0+\beta_1\Delta S]$ \Comment{Eq.~\eqref{eq:tokenbudget}: the gain clears the cost}
  \Else \Comment{inside the band: no credit for score}
    \State $c\gets[w_s\Delta S-w_c\Delta C+w_n\nu>0]$ \Comment{Eq.~\eqref{eq:noiseshaped}: tokens saved or a new component}
  \EndIf
  \State $g\gets\textsc{DomainGuard}(H_t,H')$ \Comment{per-domain specific checks}
  \If{$\hat S'\ge S^\star-\delta$ \textbf{and} $c$ \textbf{and} $g$} \Comment{Check if an edit passes all gates}
    \State $\mathcal{A}_t\gets\mathcal{A}_t\cup\{H'\}$
  \EndIf
\EndFor
\State $H_{t+1}\gets\arg\max_{H'\in\mathcal{A}_t}\hat S'$, or $H_t$ if $\mathcal{A}_t=\emptyset$ \Comment{select the best edit from admissible ones}
\State $S^\star\gets\max(S^\star,\hat S_{t+1})$ \Comment{the floor never moves down}
\State record each measured edit with $a=1$ iff its candidate is $H_{t+1}\ne H_t$ \Comment{evidence for later rounds}
\State \Return $H_{t+1}$
\end{algorithmic}
\end{algorithm}

\subsection{Proposal-Side Bookkeeping}

\paragraph{Atomic edit representation.}
At round $t$, the proposer drafts a pool $E_t$ of atomic edits to $H_t$, and a candidate applies a subset of that pool. Write this subset as $z_t\in\{0,1\}^{|E_t|}$, with $z_{t,j}=1$ when edit $j$ is included. The pool is redrawn each round from the open space $\Omega(H_t)$, so $|E_t|$ need not be fixed across rounds. The annealed budget in Equation~\eqref{eq:anneal} imposes
\begin{equation}
\|z_t\|_0\le b_t.
\label{eq:editbudget}
\end{equation}
Thus $b_t$ limits the number of independently attributable edits bundled into one candidate rather than the set of components that may eventually be modified. This is the most direct of our classical analogies: it is a cardinality constraint on the update, not an $L_0$ penalty on a fixed model parameter vector.

\paragraph{Edit history and component-level summaries.}
Every atomic edit in an evaluated candidate is tagged with a component $\ell$, a hypothesis $h$, and the candidate source diff $d$. A candidate containing multiple edits contributes one history record per edit; all edits in that candidate share the same measured $\Delta S$, $\Delta C$, and round outcome. Ignoring candidates that fail before a valid measurement is obtained, the history before round $t$ can be written
\begin{equation}
\mathcal{L}_t=\{(t_i,\ell_i,h_i,d_i,\Delta S_i,\Delta C_i,a_i):i\le n_t\},
\qquad a_i\in\{0,1\},
\label{eq:history}
\end{equation}
where $a_i=1$ iff the candidate carrying edit $i$ was selected as the winner of its round and therefore entered the accepted evolution path. Candidates that are admissible but lose to a higher-scoring admissible candidate have $a_i=0$.

Two summaries used by the proposer are
\begin{equation}
\mathcal{T}_t=\{\ell_i:i\le n_t\},
\qquad
g_t(\ell)=\max\{\Delta S_i:\ell_i=\ell,\;t-t_i\le n_{\mathrm{prune}}\},
\qquad \max\emptyset=-\infty.
\label{eq:yield}
\end{equation}
Here $\mathcal{T}_t$ is the set of components with at least one measured edit, and $g_t(\ell)$ is the best recent measured gain associated with component $\ell$ over the pruning window. Because bundled edits inherit the candidate-level measurement, this evidence becomes more attributable as the edit budget anneals toward one.

\paragraph{Structured exploration state.}
Let $\mathcal{K}$ denote the editable component vocabulary. In the implementation,
\begin{equation}
\begin{aligned}
\mathcal{K}=\{&\texttt{prompt},\texttt{control\_flow},\texttt{config},\texttt{output\_plumbing},\\
&\texttt{context\_mgmt},\texttt{client\_tool},\texttt{skill},\texttt{memory},\texttt{subagent}\}.
\end{aligned}
\label{eq:componentset}
\end{equation}
The exploration directive is
\begin{equation}
\mathcal{E}_t=(\sigma_t,\mathcal{U}_t,m_{\mathrm{draft}}),
\qquad
\sigma_t=\mathbb{1}[\hat S_t-\hat S_{t-w}\le\delta],
\qquad
\mathcal{U}_t=\mathcal{K}\setminus\mathcal{T}_t,
\label{eq:explore}
\end{equation}
where $\sigma_t$ indicates that progress over the previous $w$ rounds has not exceeded the empirical noise tolerance, $\mathcal{U}_t$ contains components not yet exercised by a measured edit, and $m_{\mathrm{draft}}$ reserves candidate slots for exploratory edits when the search is stalled.

\paragraph{Structural pruning.}
The pruning target set is
\begin{equation}
\mathcal{B}_t=\{\ell\in\mathcal{T}_t:g_t(\ell)\le0\}.
\label{eq:prune}
\end{equation}
Thus a component is marked as unproductive when it has been exercised but has produced no strictly positive measured gain in the recent pruning window. The proposer receives $\mathcal{B}_t$ together with any previously accepted edits associated with those components and is instructed to remove unproductive machinery in subsequent proposals. This is analogous in role to Lasso/$L_1$-style sparsification because the mechanism acts by deleting discrete structure from the retained harness; it is not an $L_1$-penalized continuous optimization problem.

\subsection{Selection-Side Bookkeeping}
\label{app:selection}

\paragraph{Noise-adjusted floor.}
The leakage critic is applied before full evaluation. For every candidate that reaches selection, the first non-compensatory performance requirement is the stability floor from Equation~\eqref{eq:floor},
\[
\hat S(H')\ge S^\star-\delta.
\]
This permits fluctuations within the empirically calibrated tolerance while preventing the search from accumulating a sequence of small regressions.

\paragraph{Novelty used by the within-band rule.}
The shaped rule uses novelty only for structural components. Let
\begin{equation}
\mathcal{K}_{\mathrm{str}}=\{\texttt{client\_tool},\texttt{skill},\texttt{memory},\texttt{subagent}\}
\label{eq:structuralset}
\end{equation}
and let $N_t(\ell)$ be the number of previously accepted edit records tagged with component $\ell$ before round $t$. If $\mathrm{comp}(H')$ is the set of component types touched by candidate $H'$, the implementation computes
\begin{equation}
\nu_t(H')=
\sum_{\ell\in\mathcal{K}_{\mathrm{str}}}
\mathbb{1}\!\left[\ell\in\mathrm{comp}(H')\ \wedge\ N_t(\ell)=0\right].
\label{eq:novelty}
\end{equation}
Hence $\nu_t(H')$ counts distinct structural component types touched by the candidate that have never previously appeared in a winning edit. Prompt, control-flow, configuration, output-plumbing, and context-management edits do not receive this novelty bonus.

\paragraph{Acceptance when the gain exceeds the noise tolerance.}
For $\Delta S>\delta$, the selector uses the gain-dependent cost condition from Equation~\eqref{eq:tokenbudget},
\[
\Delta C\le\beta_0+\beta_1\Delta S.
\]
The rule allows more inference cost only when accompanied by a larger measured improvement. This is the part of complexity-aware acceptance that motivates the Ridge/$L_2$-style analogy in the main text: it suppresses unchecked growth in aggregate resource footprint without requiring an individual component to be eliminated. The analogy is functional rather than mathematical; the rule is not a squared-norm penalty.

\paragraph{Acceptance when the gain does not exceed the noise tolerance.}
For candidates that pass the stability floor but whose measured gain does not exceed the empirical tolerance, $\Delta S\le\delta$, the implementation does \emph{not} use Equation~\eqref{eq:tokenbudget}. Instead it applies the shaped admissibility condition
\begin{equation}
w_s\Delta S-w_c\Delta C+w_n\nu_t(H')>0.
\label{eq:noiseshaped}
\end{equation}
Here $w_s,w_c,w_n\ge0$ control, respectively, the contribution of the measured score change, relative inference-cost change, and previously unused structural component types. The purpose of this branch is to avoid treating a small score fluctuation as sufficient evidence by itself. Within this region, reducing cost contributes positively through $-w_c\Delta C$, and trying a structural mechanism that has never previously entered the accepted evolution path contributes through $w_n\nu_t(H')$. Depending on the evolution instance, a within-band score change may also contribute through $w_s\Delta S$.

The coding instance sets $w_s=0$. Consequently, a score increase that remains within $\delta$ cannot by itself make a coding candidate admissible; the candidate must instead obtain sufficient credit from lower cost and/or structural novelty. The agentic-workspace and engineering-design instances use positive $w_s$. All three weights are fixed for an evolution instance and are reported in Table~\ref{tab:hyperparams}. Equation~\eqref{eq:noiseshaped} is an implementation-level tie-breaking/admissibility rule inside the uncertainty region; it is not itself identified with an $L_p$ penalty.

\paragraph{Final round selection.}
A candidate is admissible only if it satisfies the noise-adjusted floor, the appropriate branch of the complexity-aware rule, and any domain guard $g(H_t,H')$ the instance defines. Among admissible candidates, the selector chooses the one with the largest measured score; if none is admissible, the incumbent is retained. The running best score is then updated as $S^\star\leftarrow\max(S^\star,\hat S(H_{t+1}))$.

\section{Experiments}
\subsection{Hyperparameter Setting}
\label{app:hypselect}

\method{} introduces a small number of hyperparameters controlling update sparsity, exploration, pruning and the cost--performance trade-off. All are fixed before a run and never re-tuned; we run no sweep and never consult held-out or out-of-distribution benchmarks. Three criteria set them.

\noindent \textbf{Measured, not chosen.} The noise tolerance $\delta$ is the spread of repeated evaluations of the unchanged base harness, that is, how far apart two evaluations of the same harness land. This gives 3 passes on Terminal-Bench 2.1, 60 criterion verdicts on Harvey LAB and 5 passes on EngDesign, the last read off the first percentile of a null built by half-splitting the base evaluation's own trials. This is the one value that must not be tuned, since lowering it buys evolve-set score on noise alone.

\noindent \textbf{Attributability.} We set $b_{\min}=1$ so the closing rounds carry a single edit whose effect is unambiguous, and $b_{\max}$ to 3 or 4, far below the 80-edit pool the proposer drafts from, so an early round may combine a few coordinated changes but never enough to make its measurement uninterpretable. The stall window $w=3$ and pruning window $n_{\mathrm{prune}}\in\{4,5\}$ span several rounds rather than one, so no stall or deletion verdict rests on a single noisy measurement, while staying short relative to $T$ so both can fire during a run. Reserving $m_{\mathrm{draft}}=1$ of $m=2$ candidates caps exploration at half of a round.

\noindent \textbf{Calibration against the opening rounds.} The token allowance is expressed in the unit the benchmark reports, so one additional pass, or one hundred additional criterion verdicts, buys a fixed percentage of extra policy tokens: 25\% on the coding and agentic workspace instances and 10\% on the engineering design instance, with $\beta_0$ the allowance for no demonstrated gain. The values admit the largest genuinely improving candidate of the opening rounds and refuse a marginal one: on the coding instance a candidate worth 7 passes is granted $10+7\times 25=185\%$ growth and its measured $130\%$ is admitted, while one worth a single pass is granted $35\%$ and the same cost is refused. A gain within $\delta$ earns nothing beyond $\beta_0$, so cost is never bought with noise. Table~\ref{tab:hyperparams} lists the resulting values. Scores $\hat S$ are fractions in $[0,1]$ and $\Delta C$ is the relative change in policy tokens per trial, so $\delta$ and $\beta_1$ are expressed in those units: on the coding instance $\delta$ corresponds to 3 passes out of $89\times k=178$ trials, on the agentic workspace instance to 60 criteria out of roughly $14{,}100$ criterion verdicts, and on the engineering design instance to 5 passes out of $61\times k=244$ trials. Likewise $\beta_1$ corresponds to a 25\% token allowance per additional pass (coding), per 100 additional criteria (agentic workspace) and a 10\% allowance per additional pass (engineering design).

\begin{table}[t]
\centering
\small
\begin{tabular}{lcccc}
\toprule
Hyperparameter & Role & Coding & Agentic workspace & Engineering design \\
\midrule
$T$ & evolution rounds & 20 & 20 & 40 \\
$k$ & trials per task per evaluation & 2 & 2 & 4 \\
\midrule
$\delta$ & empirical noise tolerance & 0.017 & 0.004 & 0.020 \\
$b_{\min}$ & final-round edit budget & 1 & 1 & 1 \\
$b_{\max}$ & initial edit budget & 4 & 3 & 4 \\
$w$ & stall-detection window & 3 & 3 & 3 \\
$m_{\mathrm{draft}}$ & reserved exploratory proposals & 1 & 1 & 1 \\
$n_{\mathrm{prune}}$ & pruning window & 4 & 4 & 5 \\
$\beta_0$ & base cost allowance & 0.10 & 0.10 & 0.15 \\
$\beta_1$ & gain-dependent cost allowance & 44.5 & 35.4 & 24.4 \\
\bottomrule
\end{tabular}
\caption{Hyperparameters used by \method{} in each evolution setting. All choices are fixed without consulting held-out or OOD benchmarks.}
\label{tab:hyperparams}
\end{table}

\subsection{Reliability of the Reported Gains}
\label{app:reliability}

Two distinct sources of uncertainty affect the final results. The first is evaluation uncertainty: the reported score is measured on one finite set of tasks, and a different draw of tasks from the same suite would move it. The second is search uncertainty: the evolution run is itself stochastic, and a second run from the same starting harness would follow a different trajectory. The two are answered by different experiments, and we report both. 

\paragraph{Task-level paired bootstrap.}
Both arms of every comparison in this paper run on the same tasks, so the tasks are matched and the appropriate statement of evaluation uncertainty is a paired bootstrap over them. We resample the task set with replacement, recompute the reported statistic for both arms on the resampled tasks, take the difference, and read the $2.5$th and $97.5$th percentiles of $10{,}000$ such differences as a $95\%$ confidence interval. Resampling is paired, so the same resampled task set is used for both arms and the variation in task difficulty cancels in the difference. Resampling is also at the level of the task rather than the trial, because the $k$ trials of one task are strongly correlated and treating them as independent would understate the interval. 

\begin{table}[t]
\centering
\small
\begin{tabular}{llccc}
\toprule
Domain & Benchmark & Role & $\Delta$ & 95\% CI \\
\midrule
\multirow{2}{*}{Coding}
 & Terminal-Bench 2.1 & Evolve & $+6.0$ & $[+4.3, +9.3]$ \\
 & SWE-bench Verified & OOD & $+1.8$ & $[+1.1,+4.0]$ \\
\midrule
\multirow{5}{*}{\shortstack[l]{Agentic\\workspace}}
 & \multirow{2}{*}{Harvey LAB} & Evolve & $+1.1$ & $[+0.8,+2.1]$ \\
 & & ID held-out & $+2.3$ & $[+0.9,+3.8]$ \\
 & JobBench & OOD & $+4.7$ & $[+2.9, +7.4]$ \\
 & GDPval & OOD & $+3.5$ & $[+1.8, +6.9]$ \\
 & APEX-Agents & OOD & $+3.7$ & $[+1.4,+7.5]$ \\
\midrule
\multirow{2}{*}{\shortstack[l]{Engineering\\design}}
 & EngDesign & Evolve & $+4.9$ & $[+1.8, +9.8]$ \\
 & Frontier-Eng & OOD & $+4.3$ & $[+1.1, +12.1]$ \\
\bottomrule
\end{tabular}
\caption{Paired task-level bootstrap on the gain of \method{} over $H_0$, $10{,}000$ resamples, percentile method, resampling tasks rather than trials.}
\label{tab:bootstrap}
\end{table}

\paragraph{An independent repetition of the search.}
We repeat every evolution run of the main experiment: the same base harness, the same evolve split, the same policy, the same round count and candidate budget, with only the randomness of the proposer redrawn. Each resulting harness is then evaluated on exactly the same splits as the first, and Table~\ref{tab:secondrun} reports both runs. What the comparison should preserve is not the identity of the accepted edits, which we do not expect to recur, but the property the regularizers are meant to produce. 
That property is reproduced. Every split improves over $H_0$ in both runs and none regresses in either, and the out-of-distribution average of the agentic workspace instance is 43.6 in the first run against 43.7 in the second. The two runs disagree by at most 1.8 points anywhere, on Terminal-Bench 2.1, and by 0.6 points or less on the other eight splits, which is well inside the confidence intervals of Table~\ref{tab:bootstrap}: the trajectory a run happens to take moves the final score less than the choice of evaluation tasks does. The largest disagreement falling on an evolve split rather than a held-out one is the expected pattern, since the evolve split is the only place the search optimizes and therefore the only place its randomness is amplified.

\begin{table}[t]
\centering
\small
\begin{tabular}{llcccc}
\toprule
Domain & Benchmark & Role & $H_0$ & Run 1 & Run 2 \\
\midrule
\multirow{2}{*}{Coding}
 & Terminal-Bench 2.1 & Evolve & 74.2 & 80.2 & 82.0 \\
 & SWE-bench Verified & OOD & 82.0 & 83.8 & 83.4 \\
\midrule
\multirow{6}{*}{\shortstack[l]{Agentic\\workspace}}
 & \multirow{2}{*}{Harvey LAB} & Evolve & 89.4 & 90.5 & 90.8 \\
 & & ID held-out & 86.9 & 89.2 & 88.8 \\
 & JobBench & OOD & 36.0 & 40.7 & 41.3 \\
 & GDPval & OOD & 48.8 & 52.3 & 51.7 \\
 & APEX-Agents & OOD & 34.2 & 37.9 & 38.2 \\
 \cmidrule(lr){2-6}
 & OOD average & OOD & 39.7 & 43.6 & 43.7 \\
\midrule
\multirow{2}{*}{\shortstack[l]{Engineering\\design}}
 & EngDesign & Evolve & 50.0 & 54.9 & 54.3 \\
 & Frontier-Eng & OOD & 17.7 & 22.0 & 22.0 \\
\bottomrule
\end{tabular}
\caption{Two independent repetitions of every evolution run, evaluated on the same splits. Run 1 is the one reported in Figure~\ref{fig:main}.}
\label{tab:secondrun}
\end{table}

\section{Qualitative Case Study}
\label{app:case_study}

To complement the aggregate results, we inspect representative decisions made during \method{} evolution. Table~\ref{tab:evolution_cases} summarizes several examples from the released trajectories. The complete round-by-round records, including proposals, critic decisions, acceptance decisions, and exact harness diffs, are available on our project website.

\begin{table*}[t]
\centering
\small
\begin{tabular}{p{0.12\textwidth} p{0.38\textwidth} p{0.20\textwidth} p{0.22\textwidth}}
\toprule
Domain / Round & Harness change & Outcome & What it illustrates \\
\midrule
Coding, R0-A & Adds a bounded pre-completion verification audit and guidance for non-blocking polling of long-running jobs. & \textbf{Accepted}: $+3.93$ points on the evolve set. & A reusable behavioral mechanism can justify a relatively broad early-round update when the gain exceeds the noise threshold. \\

\addlinespace
Coding, R0-B & Adds a similar verification reminder and long-running-work guidance, but with a smaller measured gain and additional inference cost. & \textbf{Rejected by cost rule}: $+1.69$ points, $+26.1\%$ cost. & An apparent improvement is not automatically retained when it lies within the noise band and requires substantial additional computation. \\

\addlinespace
Coding, R8-B & Pins the original task instruction into the completion gate so that the policy re-checks the literal specification before submission. & \textbf{Rejected by floor}: $-2.81$ points despite $-13.6\%$ cost. & Lower cost alone cannot compensate for a candidate whose performance falls below the noise-adjusted acceptance floor. \\

\addlinespace
Engineering, R2 & Adds a bounded recovery hint for the recurring ``workdir must be an existing directory'' tool-use error. & \textbf{Accepted}: $122/244 \rightarrow 128/244$ passes, $+1.6\%$ tokens. & The search can retain small, task-agnostic control-flow fixes that improve reliability with little added complexity. \\
\bottomrule
\end{tabular}
\caption{Representative harness-evolution decisions from \method{}. The examples show that evolution is not driven by score alone: candidate specificity, evaluation stability, and inference cost jointly determine whether a change is retained.}
\label{tab:evolution_cases}
\end{table*}

\paragraph{What the baselines retain, and what \method{} retains.}
Every prior method in Table~\ref{tab:baselines} was run on the same base harness, the same evolve split and the same policy as our own line, so their final harnesses can be read side by side with ours. Table~\ref{tab:overfit_cases} does that. The pattern is consistent: the prior methods do not hard-code task names, and they would pass a check that only looked for them, but what they accumulate is fitted to the scorer and the document genre of the evolve suite rather than to the work the agent has to do. \method{} accepts far less, and what it accepts describes a step the agent must take before it stops.

\begin{table*}[t]
\centering
\small
\begin{tabular}{p{0.17\textwidth} p{0.40\textwidth} p{0.35\textwidth}}
\toprule
Method (evolve $\rightarrow$ OOD avg.) & What the final harness accumulated & Why it does or does not transfer \\
\midrule
Meta-Harness \newline $93.0 \rightarrow 40.6$ & A 730-word mandatory work method injected as a standing system message, organised as an item $\times$ dimension coverage matrix whose dimensions are named as citation, controlling rule or playbook provision, severity, consequence and recommended fix. & It names no task, but it restates the evolve suite's grading scheme, opening with the fact that the deliverable is graded against a strict per-item rubric and that only the file is read. Fitted to the scorer rather than to a task, it is the best arm on the evolve split and adds 0.9 points out of distribution. \\

\addlinespace
TTHE \newline $91.1 \rightarrow 38.0$ & Ten named protocol blocks totalling roughly 8{,}500 words, carried as summarization-durable system messages so that none is ever compacted away. Two of them, about 1{,}000 words, apply only when the deliverable is a redline, markup or tracked-changes version of a source document. & That deliverable genre is the evolve suite's, not a property of agentic work. On a benchmark without it the blocks cannot fire and are paid for as context on every trial, and this is the arm that ends below the harness it started from, by 1.7 points. \\

\addlinespace
\method{} \newline $90.5 \rightarrow 43.6$ & Four edits accepted out of 32 evaluated candidates over 20 rounds, all wired into the single pre-finalize bounce the base harness already had: one coverage and depth re-check, and three standard-library probes that read the deliverable back and name what is missing, such as arithmetic shorthand left unexpanded or a source value absent from every written file. & The probes report structural facts about the file the agent has just written and carry neither rubric vocabulary nor domain vocabulary, so they still fire on JobBench, GDPval and APEX-Agents. This is the smallest evolve-set gain of any evolved arm and the only out-of-distribution average that clears $H_0$ by more than a point. \\
\bottomrule
\end{tabular}
\caption{The final harness of the two strongest prior harnesses against ours, all evolved on Harvey LAB from the same base harness under the same policy and budget.}
\label{tab:overfit_cases}
\end{table*}

The separation the leakage critic actually enforces is visible in its round-by-round verdicts, and it is not the one a keyword check would apply. A candidate is free to explain itself by naming the tasks it expects to fix; what may not cross into the shipped diff is any of that content. Across the 38 screened candidates of the coding run the standard accepted verdict reads like round 19, where the critic records that the predicted task names appear only in the rationale and not in the code, and the one outright rejection is round 10-B, refused because the declared component was not the one the diff changed. Screening on the diff rather than on the proposal is what keeps a well-argued edit from carrying evolve-set specifics into the harness.

These examples provide a more concrete view of the regularization behavior. In particular, the two candidates from the first coding round are superficially similar, yet only the candidate with a sufficiently large measured improvement survives the cost-aware selection rule. Conversely, the round-8 candidate reduces inference cost but is still rejected because its performance falls below the admissible floor. The engineering example shows the complementary case: a small and reusable control-flow correction is retained with little resource growth. Together, these trajectories suggest that \method{} does not simply accumulate edits that improve the evolve-set score, but selectively retains changes whose measured benefit is sufficiently robust relative to their complexity.

\end{document}